# Performance Evaluation of Predictive Classifiers For Knowledge Discovery From Engineering Materials Data Sets

Doreswamy, Hemanth. K. S

*Abstract*—In this paper, naive Bayesian and C4.5 Decision Tree Classifiers(DTC) are successively applied in materials informatics to classify the engineering materials into different classes for the selection of materials that suit the input design specifications. Here, the classifiers are analyzed individually and their performance evaluation is analyzed with confusion matrix predictive parameters and standard measures, the classification results are analyzed on different class of materials. Comparison of classifiers has found that naive Bayesian classifier is more accurate and better than the C4.5 DTC. The knowledge discovered by the naive Bayesian classifier can be employed for decision making in materials selection in manufacturing industries.

## I. INTRODUCTION

Engineering materials are the artificial materials that play major role in building construction, roads, bridges, irrigation systems, pipelines, machines, transportation equipment, electricity systems, tools, furniture, communication facilities, instrumentation, and various utilities and appliances both at home and in the office. There are more than one lakh materials available for use in engineering applications. Most materials fall into either one of three classes- polymer, ceramic and metal. These classifications are based on their atomic bonding forces of a particular material [28]. Materials scientists and engineers now have a rapidly evolving ability to tailor materials from the atomic scale upwards to obtain desired properties[20],[28]. Additionally, different materials can be combined to create a composite material. Within each of these classifications, materials are often further organized into groups based on their chemical composition or certain physical or mechanical properties. As a result, designing of composite materials, the storage and organization of these materials data sets are challenging tasks and are being faced by most of the manufacturing industries. Hence materials informatics is emerged as a new research discipline for employing information technology to materials science and technology.

The data mining and materials informatics are combined in this paper to discover knowledge from the engineering materials datasets.

The rest of the paper is organized as follows. The section 2 describes the scope of Data Mining on materials informatics. Naive Bayesian classifier and C4.5 Decision Tree classifier algorithms are discussed in section 3. The performance evaluation and experimental results are presented in the section 4. The conclusions and future scope are given in the section 5.

## II. SCOPE OF DATA MINING ON MATERIALS INFORMATICS

In the past two decades, an extensive research and development activities were done on scientific knowledge discovery systems. Some research and development activities are being progressed in various universities and research institutions in India[7],[15] and abroad on of materials and their performance evaluation [6],[24],[26],[27]. At present, the development of materials science and technology and with the continuous emergence of new materials, manufacturing technologies, processes and methods, and proliferation of materials data sets have been increased enormously. Further, difficulties encountered by researchers were how to manage and store various materials data efficiently, how to dig out valuable data to a great extent, and how to share the resources of material data and realize knowledge innovation when appealing in material science researches[27]. The quick development of information technology [26] had provided the solution of these existing problems. With the combination of material science and information science for the development of materials design, materials informatics, a new discipline, has emerged and developed quickly. Therefore, materials informatics is a cross or edge discipline with the combination and permutation of material science and information science.

Systematic design and development of materials informatics [26],[27] is inspired by researchers, technical experts and composite materials designers as fundamental engineering materials data and information are indispensable. Such data is used for materials design and simulations and also for selecting materials for equipment design and optimum use of materials. There are two ways of gathering physical, chemical and engineering fundamental

---
**Doreswamy** is with the Department of Post Graduate Studies and Research in Computer Science, Mangalore University,Mangalagangotri-574 199, Karnataka, INDIA,
e-mail: doreswamyh@yahoo.com
**Hemanth K S** is with the Department of Post Graduate Studies and Research in Computer Science, Mangalore University,Mangalagangotri-574 199, Karnataka, INDIA,
e-mail: reachhemanthmca@gmail.com

data. One is to collect brochures of materials' manufacturers, make a database of the data, and distribute it. The other is to accumulate measured data obtained by various research institutes by carrying out materials tests or data collected from scientific reference materials and compile them into a database. Therefore, a database containing characteristic values as well as information on the manufacturing process of a material, measuring equipment, shape, size and test conditions of a test sample, and test organizations needs to be constructed and organized to enable physical, chemical and engineering fundamental data to be used in industry. However, there is no research institute in the world that collects data systematically, compiles databases and makes the databases widely available. Concerning such database-making activity, it is important to construct databases efficiently through international tie-ups while avoiding the duplication of similar work, and to establish databases as public goods to be used all over the world. Creation of centralized materials database/Data warehouse[17],[24] has potential research scope in materials informatics[26] for Data Mining and Knowledge discovery for advanced engineering materials design applications.

Data Mining[16],[21] is a process of extracting previously unknown and potentially useful knowledge from the large volume of data sets. It has emerged as a dominant tool in a broad range of applications[22],[27]. Data miners consider about problems with a fairly different focus than conventional scientists, and Data Mining techniques present the possibility of making quantitative predictions in many areas where conventional approaches have had limited success. Scientists usually try to make predictions during constitutive relations [22] derived mathematically from basic laws of physics, such as the diffusion equation or the ideal gas law. However, in many areas, including materials design and development, the problems are so complex that constitutive relations either cannot be derived, or too approximate or intractable for practical quantitative use[6]. The viewpoint of a Data Mining approach [4] is to assume that useful constitutive relations exist, and to attempt to derive them primarily from data, rather than from basic laws of physics. However, today's data sets can be many orders of magnitude larger, and impressive arrays of computational algorithms[22][,[23] have been developed to computerize the task of identifying relationships within data.

Data Mining is becoming an increasingly valuable tool in the broad area of materials development and design[4],[9], and there are good reasons why this area is particularly rich for Data Mining applications. There is a massive range of possible new materials, and it is often complex to physically model the relationships between constituents, processing and final properties. Therefore, materials are primarily still developed by trial-and-error[4 ],[29], where researchers are guided by experience and heuristic rules for materials classification and property predictions are applied to somewhat limited materials data sets of constituents and processing conditions, but then try as many combinations as possible to find materials with desired properties. This is essentially human Data Mining, where one's brain, rather than the computer, is being used to find correlations, make predictions, and design optimal strategies. Transferring Data Mining tasks from human to computer offers the potential to enhance accuracy, handle more data, and to allow wider dissemination of accrued knowledge [9],[13],[14],[15 ],[16],[23].

III. DATA MING TECHNIQUE

A.  *Naive Bayesian Classifier*

Naive Bayesian classifier is a statistical method that can predict class membership probabilities such as the probability that a given tuple belongs to a particular class. Bayesian classifier is based on the Bayes theorem and it assumes that the effect of an attribute value on a given class is independent of the values of the other attributes. This assumption is called class conditional independence. It is made to simplify the computations involved and in this sense, is called as "naive".

In simple terms, a naive Bayesian classifier assumes that the presence (or absence) of a particular feature of a class is unrelated to the presence (or absence) of any other feature. For example, a fruit may be considered to be an apple if it is red, round, and about 4" in diameter. Even though these features depend on the existence of the other features, a naive Bayesian classifier considers all of these properties to independently contribute to the probability that this fruit is an apple.

Depending on the precise nature of the probability model, naive Bayesian classifier can be trained very efficiently in a supervised learning setting. In many practical applications, parameter estimation for naive Bayes models uses the method of maximum likelihood; in other words, one can work with the naive Bayes model without believing in Bayesian probability or using any Bayesian methods. In spite of their naive design and apparently oversimplified assumptions, naive Bayes classifiers often work much better in many complex real-world situations than one might expect. Recently, careful analysis of the Bayesian classification problem has shown that there are some theoretical reasons for the apparently unreasonable efficacy of naive Bayesian classifiers. An advantage of the naive Bayesian classifier is that it requires a small amount of training data to estimate the parameters (means and variances of the variables) necessary for classification. Because independent variables are assumed, only the variances of the variables for each class need to be determined and not the entire covariance matrix.

The naive Bayesian classifier is fast and incremental can deal with discrete and continuous attributes, has excellent performance in real-life problems. In this paper, the algorithm of the naive Bayesian classifier is deployed successively enabling it to solve classification problems while retaining all advantages of naive Bayesian classifier. The comparison of performance in various domains of materials classes confirms the advantages of successive learning and suggests its application to other learning algorithms.

*Algorithm of naive Bayesian Classifier*

The naive Bayesian classifier, or simple Bayesian classifier generally used for classification or prediction task. As it is simple, robust and generality, this procedure is deployed for various applications such as materials damage detection[1],[2], agricultural land soils classification[3], web page classification[5], machine learning applications[19]. Therefore the application of this is extended to classification of engineering materials data sets[27] and to reduce the computational cost involved in materials classification and selection process.

The naive Bayesian algorithm [16] is as follows:

**Input :**
- Training Data Set D with their associated class labels.

**Output:** Classification of Classes.

Method

1. Training Set D., Initialize X with one component.

2. If $P(C_i/X) > P(C_j/X)$ for all $1 \leq j \leq m; j \neq i$.
   Maximize $P(C_i/X)$

3. Compute $P(C_i/X) = \dfrac{P(X/C_i)P(C_i)}{P(X)}$

4. $P(X/C_i)P(C_i)$ need be maximized.

5. $P(X/C) = \prod\limits_{k=1}^{n} P(X_k/C_i)$
   $= P(X_1/C_i) \times P(X_2/C_i) \times P(X_2/C_i) \times \ldots \ldots P(X_n/C_i)$
   value of attribute $A_k$, for tuple X

6. if ($A_k$ = categorical) then $P(X/C_i)/P(C_i)$
   else $P(X/C_i) = g\left(X_k, \mu_{C_i}, \sigma_{C_i}\right)$

7. To predicate the class label X, $P(X/C_i)P(C_i)$
   $P(X/C_i)P(C_i) > P(X/C_j)P(C_j)$ for all $1 \leq j \leq m; j \neq i$

8. Output the classifier.

### B. C4.5 Decision Tree Classifier

Decision trees are one of the most popular methods used for inductive inference. they are robust for the noisy data and capable of leaning disjunctive expressions. A decision tree is a machine learning approach, where each internal node specifies a test on some attributes k-ary from the input feature set used to represent the data. decision tree learning has been successfully used in many practical application such as MARVIN,BACON and INDUCE[4]. These applications become classical examples of the successful use of decision trees for classification tasks. Decision tree learning is widely used because of its high-level of robustness , good performance with large data in a short time, and simple visualization and interpretation.

*Algorithm of C4.5 Decision Tree Classifier*

The C4.5 Decision Tree Classifier algorithm starts with the entire set of tuples in the training set, selects the best attribute that yields maximum information for classification, and generates a test node for the attribute. then, top down induction of decision trees divides the current set attribute. classifier generation stops, if all the tuples in a subset belong to the same class[16].

**Input:**
- Data partition, D, which is a set of training tuples and their associated class labels;
- attribute list, the set of candidate attributes;
- Attribute selection method, a procedure to determine the splitting criterion that "best" partitions the data tuples into individual classes. This criterion consists of a splitting attribute and, possibly, either a split point or splitting subset.

**Output:** A decision tree.

Method:
1. create a node N;
2. if tuples in D are all of the same class, C then
3. return N as a leaf node labeled with the class C;
4. if attribute list is empty then
5. return N as a leaf node labeled with the majority class in D; // majority voting
6. apply Attribute selection method(D, attribute list) to find the "best" splitting criterion;
7. label node N with splitting criterion;
8. if splitting attribute is discrete-valued and multiway splits allowed then // not restricted to binary trees
9. attribute list attribute list // splitting attribute; // remove splitting attribute
10. for each outcome j of splitting criterion // partition the tuples and grow subtrees for each partition
11. let Dj be the set of data tuples in D satisfying outcome j; // a partition
12. if Dj is empty then
13. attach a leaf labeled with the majority class in D to node N;
14. else attach the node returned by Generate decision tree (Dj, attribute list) to node N; endfor
15. return N;

The C4.5 algorithm usually uses an entropy-based measure know as information gain as a heuristic for the selecting the attribute that will best split the training data into separate classes. its algorithm computes the information gain of each attribute, and each attribute, and in each round, the one with the highest information gain will be chosen as the test attribute for the given set of training data. A well-chosen split point should help in splitting the data to the best possible extent. After all, a main criterion in the greedy decision tree approach is to build shorter trees. The best split point can be easily evaluated by considering each unique value for that feature in the given data as a possible split point and calculating the associated information gain.

*Information Gain*

The critical step in decision trees is the selection of the best test attribute. The information gain measure is used to select the test attribute at each node in the tree.

Let node N represent or hold the tuples of partition D. The attribute with the highest information gain is chosen as the splitting attribute for node N. This attribute minimizes the information needed to classify the tuples in the resulting partitions and reflects the least randomness or "impurity" in these partitions. Such an approach minimizes the expected number of tests needed to classify a given tuple and guarantees that a simple (but not necessarily the simplest) tree is found.

$$\text{Info}(D) = -\sum_{i=1}^{m} P_i \log_2(P_i) \quad (1)$$

Where $P_i$ is the probability that an arbitrary tuple in the D belongs to class $C_i$ and estimated by $|C_{i,D}|/|D|$. Now, partition the tuples in D on some attribute A having v distinct values, $\{a_1, a_2, ....., a_v\}$, as observed from the training data. If A is discrete-valued, these values correspond directly to the v outcomes of a test on A. Attribute A can be used to split D into v partitions or subsets, $\{D_1, D_2, ....., D_v\}$, where Dj contains those tuples in D that have outcome $a_j$ of A. is given by

$$\text{Info}_A(D) = \sum_{j=1}^{V} \frac{|D_j|}{|D|} * \text{Info}(D_j) \quad (2)$$

the term $\frac{|D_j|}{|D|}$ acts as the weight of the j$^{th}$ subset. Info$_A$(D) is the expected information required to classify a tuples from D based on the subset by A.

Information gain is defined as the difference between the original information requirement and the new requirement. i.e.

$$\text{Gain}(A) = \text{Info}(D) - \text{Info}_A(D) \quad (3)$$

The attribute with highest information gain is chosen as the test attribute for the current node. Such approach minimizes the expected number of tests needed to classify an object and guarantees that a simple tree is found.

*Gain Ratio*

The information gain measure is based towards tests with many outcomes. that is, it prefers to select attribute having a large number of values.

C4.5, a successor of ID3, uses an extension to information gain known as gain ratio, which attempts to overcome this bias. It applies a kind of normalization to information gain using a "split information" value defined analogously with Info(D) as

$$\text{SplitInfo}_A(D) = -\sum_{j=1}^{V} \frac{|D_j|}{|D|} * \log_2\left(\frac{|D_j|}{|D|}\right) \quad (4)$$

And the Gain ratio is

$$\text{GainRatio}(A) = \frac{\text{Gain}(A)}{\text{SplitInfo}(A)} \quad (5)$$

By using *SplitInfo$_A$*, which is proportional to the number of values an attribute A can take, *GainRatio(A)* effectively removes the bias of information gain towards features with many values. To resolve the issue when *SplitInfo(A)* becomes very small, C4.5 lists the set of attributes with the informationGain(A) above the average information gain for that node and the it uses the gain Ratio to select the best attribute from the list.

## IV. PERFORMANCE EVALUATION

### A. Standard Metric measurements

The naive Bayesian classifier is evaluated on engineering materials data set consisting of discrete and categorical attributes. The data sets are gathered from the from [28] and the website www.matweb.com. And then these are organized as a database. The training samples attributes and their categorical values shown in the table 1 are analyzed based on the data sets depicted in[20]. The performance of the classifier on the engineering materials data sets is analyzed with confusion matrix by measuring the standard metrics that are commonly used for measuring the classification performance of other classification models.

The experiments in this research are evaluated using the standard metrics of accuracy, precision, recall and F-measure for engineering materials data set classification. These were calculated using the predictive classification table, known as Confusion Matrix[16].

Table 1:
Confusion Matrix with predictive parameters

|        |            | PREDICTED  |          |
|--------|------------|------------|----------|
|        |            | IRRELEVANT | RELEVANT |
| ACTUAL | IRRELEVANT | TN         | FP       |
|        | RELEVANT   | FN         | TP       |

Where:

TN (True Negative) : Number of correct predictions that an instance is irrelevant

FP (False Positive) : Number of incorrect predictions that an instance is relevant

FN (False Negative) : Number of incorrect predictions that an instance is irrelevant

TP (True Positive) : Number of correct predictions that an instance is relevant

Accuracy(ACC) – The proportion of the total number of predictions that were correct:

Accuracy (%) = (TN + TP)/(TN+FN+FP+TP)     (6)

Precision(PREC) – The proportion of the predicted relevant materials data sets that were correct:

Precision (%) = TP / (FP + TP)     (7)

Recall(REC) – The proportion of the relevant materials data sets that were correctly identified

Recall (%) = TP / (FN + TP)     (8)

F-Measure(FM) – Derives from precision and recall values:

F-Measure (%) = (2 x REC x PREC)/(REC + PREC)     (9)

The F-Measure was used, because despite Precision and Recall being valid metrics in their own right, one can be optimized at the expense of the other .

The F-Measure only produces a high result when Precision and Recall are both balanced, thus this is very significant.

A Receiver Operating Characteristic (ROC) analysis was also performed, as it shows the sensitivity (FN classifications) and specificity (FP classifications) of a test. The ROC analysis is a comparison of two characteristics: TPR (true positive rate) and FPR (false positive rate).

The TPR measures the number of relevant tuples that were correctly identified.

TPR = TP / (TP + FN)     (10)

The FPR measures the number of incorrect classifications of relevant tuples out of all irrelevant test tuples.

FPR = FP / (FP + TN)     (11)

### B. Experimental Results

The engineering data sets involved in naive Bayesian classification and C4.5 decision tree has 2431 data sets with twenty five attributes including numeric attributes. The categorical attributes shown in the table 2 are considered for classification. The classifier performance is tested on 3/4th training samples from the data sets. Later, class wise and whole data sets were tested for assuring the confidence of the classifier.

Here in the experiment, 2431 datasets are used in both classifiers and the performance measures considered as standard metrics- Accuracy(ACC), Precision(PREC), Recall(REC) and F-Measure(FM). The standard metrics values of naive Bayesian Classifier and C4.5 DTC are computed on confusion matrix predictive parameters respectively at class levels.

The classifiers performance on individual class are described in table 3 and table 4. The naive Bayesian classification accuracies of Polymer, Ceramic and Metal classes are respectively 93.76%, 94.61%, and 93.25%, and the C4.5 Decision Tree classification accuracies of Polymer, Ceramic and Metals are respectively 93.17%, 92.04% and 90.97%.

TABLE 2:
TRAINING DATA SETS AND THEIR ATTRIBUTES CATEGORICAL VALUES

| Attribute # | Sub Property | Polymer | Ceramics | Metals |
|---|---|---|---|---|
| 1 | CS | Poor | Excellent | Good |
| 2 | FS | Poor | Good | Good |
| 3 | CH | Poor | Poor to Fair | Good to excellent |
| 4 | CE | Nil | Poor to Good | Excellent |
| 5 | TCE | Very High | Low | High |
| 6 | W A | Poor | Poor | Poor |
| 7 | E I | Good to Excellent | Good to Excellent | Poor |
| 8 | C R | Good to Excellent | Good to Excellent | Poor to Good |
| 9 | CORR | Excellent | Good to Excellent | Very poor to Good |
| 10 | S M | Good | Poor | Excellent |
| 11 | CAST | Fair | Poor | Excellent |
| 12 | EXTRN | Good | Poor | Excellent |
| 13 | MOLD | Excellent | Fair | Good |
| 14 | MACHN | Good | Poor | Good |
| 15 | MANFT | Excellent | Good | Fair |

TABLE 3:
STANDARD METRICS VALUES OF ACCURACY, PRECISION, RECALL AND F-MEASURE OF NAIVE BAYESIAN CLASSIFIER COMPUTED ON CONFUSION MATRIX PREDICTIVE PARAMETERS

| Class Test | TP | TN | FP | FN | ACC (%) | PREC (%) | REC (%) | FM (%) |
|---|---|---|---|---|---|---|---|---|
| P | 734 | 243 | 34 | 31 | **93.76** | 95.57 | 95.95 | 95.76 |
| C | 560 | 249 | 19 | 27 | **94.61** | 96.72 | 95.4 | 96.05 |
| M | 329 | 169 | 15 | 21 | **93.25** | 95.63 | 94.00 | 94.80 |

TABLE 4:
STANDARD METRICS VALUES OF ACCURACY, PRECISION, RECALL AND F-MEASURE OF C4.5 DC CLASSIFIER COMPUTED ON CONFUSION MATRIX PREDICTIVE PARAMETERS

| Class Test | TP | TN | FP | FN | ACC (%) | PREC (%) | REC (%) | FM (%) |
|---|---|---|---|---|---|---|---|---|
| P | 731 | 239 | 36 | 35 | **93.17** | 95.30 | 95.43 | 95.36 |
| C | 553 | 245 | 19 | 41 | **92.04** | 96.68 | 93.10 | 94.85 |
| M | 321 | 163 | 19 | 29 | **90.97** | 94.41 | 91.71 | 93.04 |

Table 5 shows average standard metrics values of both the classifiers on engineering materials data set consisting of all class of materials. The average classification accuracies of naive Bayesian Classifier and C4.5 Decision Tree Classifier on engineering materials datasets is 93.95% and 91.93% respectively. This show naive Bayesian classifier can be used as the easier technique for classification with more accuracy for complex large datasets.

TABLE 5:
AVERAGE STANDARD METRIC VALUES OF NAIVE BAYESIAN AND C4.5 DT CLASSIFIERS.

| Standard metrics | Naive Bayesian(%) | Decision Tree(%) |
|---|---|---|
| Accuracy | 93.95 | 91.93 |
| Precision | 95.98 | 95.51 |
| Recall | 95.36 | 92.96 |
| F-Measure | 95.67 | 94.22 |

The ROC analysis with TPR and FPR characteristics is shown in table 6. From the ROC analysis of naive Bayesian classifier and C4.5 DTC, it is found that naive Bayesian classifier is more accurate technique than the C4.5 DTC since naive Bayesian classifier assumes that the effect of an attribute value on a given class is independent of the values of the other attributes.

TABLE 6 :
TPR (TRUE POSITIVE RATE) AND FPR (FALSE POSITIVE RATE)

|  | Naive Bayesian Classifier | C4.5 DTC |
|---|---|---|
| TPR | 0.9535 | 0.9385 |
| FPR | 0.0932 | 0.0441 |

Comparison of classification results are depicted through graphical representation as shown in Fig 1 and Fig 2.

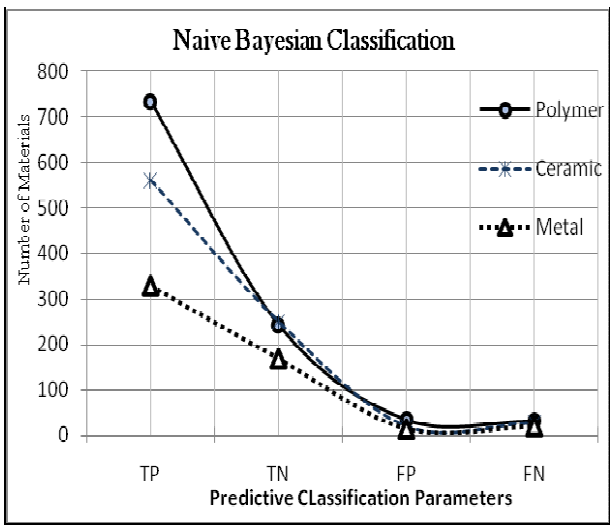

Fig 1:Classification results by Naive Bayesian Classifier confusion matrix predictive parameters

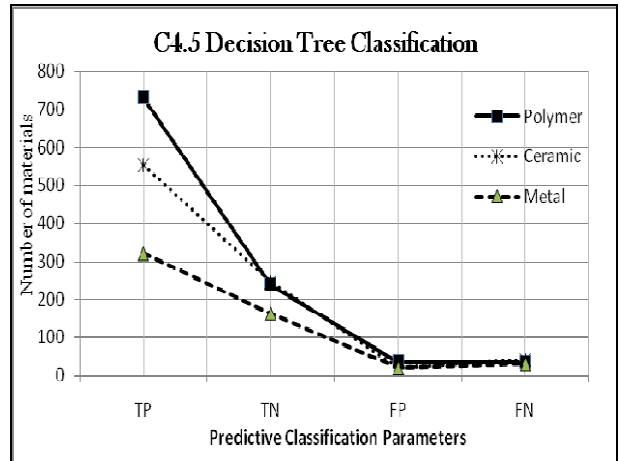

Fig 2:Classification results by C4.5 Decision tree confusion matrix predictive parameters.

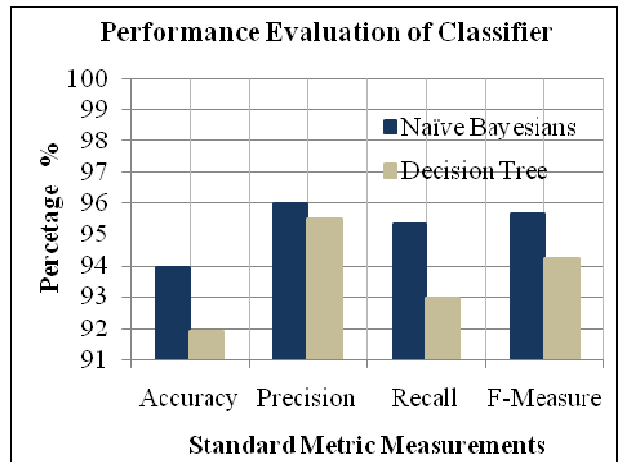

Fig 3:Performance evaluation of naive Bayesian and C4.5 Decision Tree classifiers on standard measures.

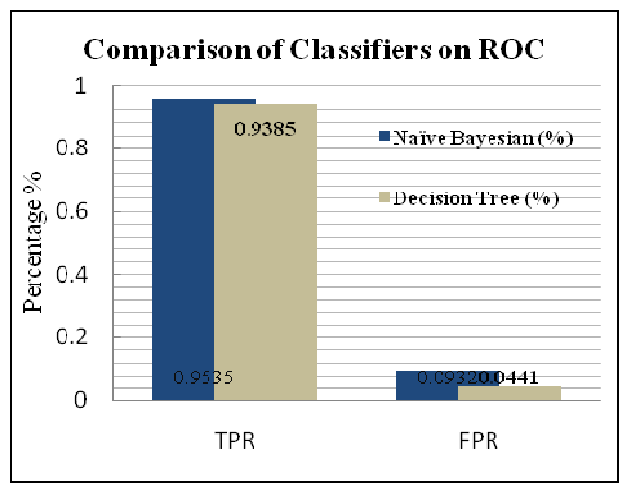

Fig 4: Receiver Operating Characteristic(ROC) Analysis

## V. CONCLUSIONS AND FUTURE SCOPE

In this paper naive Bayesian classifier and C4.5 DTC are implemented on engineering materials data sets for solving classification problem. Comparison of these classifiers are done in detail with confusion matrix predictive parameters(TP,TF,FP,FN). The performance of evaluation of these two classifiers is analyzed with standard metric measures on classifying the engineering materials data sets into polymer, ceramic and metal classes.

The classification results analyzed in this research depicts that naive Bayesian classifier is more accurate and simple technique than the C4.5 DTC. Although the results of naive Bayesian classifier are close to the result of the C4.5 DTC, both are used for predicting the materials class for the selection of materials that suit the input design specifications. Therefore, naive Bayesian classifier is proposed in materials informatics for knowledge discovery from the engineering materials data sets for advanced engineering material design applications.

Further, the classification accuracy can be improved by employing noise removal techniques for eliminating outliers in data sets.

## ACKNOWLEDGEMENT

This work has been supported by the University Grant Commission(UGC), India under Major Research Project entitled "Scientific Knowledge Discovery Systems (SKDS) For Advanced Engineering Materials Design Applications" vide reference F.No. 34-99\2008 (SR), 30th December 2008. The authors gratefully acknowledge the support of UGC and thank the unanimous reviewers for their kind comments and suggestions.

High-*Temperature Alloys and Concepts of Alloy Design for SOFC Applications*, (2002).
(30) Li-Min Wang., Xiao-Lin Li., Chun-Hong Cao ., Sen-Miao Yuan.:" Combining decision tree and Naive Bayes for Classification" *Journal of Knowledge-Based System* Vol.19.,pp.511-515(2006).
(31) Pablo D. Robles- Granda and V.Belik .:"A Comparison of Machine learning Classifies Applied to Financial Datasets" *Proceedings of the World Congress on Engineering and Computer Science* Vol 1(2010).

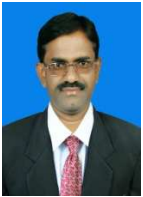
**Doreswamy**  received B.Sc degree in Computer Science and M.Sc Degree in Computer Science  from University of Mysore  in 1993 and 1995 respectively. After completion of his Post-Graduation Degree, he subsequently joined and served as Lecturer in Computer Science at St.Joseph's College, Bangalore from 1996-1999 and at Yuvaraja's College, a constituent college of University of Mysore from 1999-2002. Then he has elevated to the position Reader in Computer Science at Mangalore University in year 2003. He was the Chairman of the Department of Post-Graduate Studies and Research in Computer Science from 2003-2005 and from 2008-2009 and served at varies capacities in Mangalore University  and at present he is the Chairman of Board of Studies in Computer Science of Mangalore University. His areas of research interests include Data Mining and Knowledge Discovery, Artificial Intelligence and Expert Systems, Bioinformatics, Molecular Modeling and Simulation, Computational Intelligence, Nanotechnology, Image Processing and Pattern Recognition. He has been granted a Major Research project entitled "Scientific Knowledge Discovery Systems(SKDS) for advanced Engineering Materials Design Applications" from the funding Agency University Grant Commission, New Delhi, INDIA.  He has published about 30 contributed peer reviewed papers at National/International Journals and Conferences. He received SHIKSHA RATTAN PURSKAR  for his outstanding achievements in the year 2009 and RASTRIYA VIDYA SARAWATHI AWARD and EMINENT EDUCATIONALIST AWARD for outstanding achievement in chosen field of activity in the year 2010.

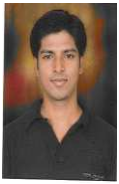
 Hemanth K S received B.Sc degree and MCA degree in the years 2006 and 2009 respectively. Currently working as Project Fellow of UGC Major Research Project  and is working towards his Ph.D degree in Computer Science under the guidance of Dr. Doreswamy  in the Department of Post-Graduate Studies and Research in Computer Science, Mangalore University.